\documentclass{article}

\usepackage{arxiv}

\usepackage[utf8]{inputenc} 
\usepackage[T1]{fontenc}    
\usepackage{hyperref}       
\usepackage{url}            
\usepackage{booktabs}       
\usepackage{amsfonts}       
\usepackage{nicefrac}       
\usepackage{microtype}      
\usepackage{lipsum}		
\usepackage{graphicx}
\usepackage{natbib}
\usepackage{doi}

\usepackage[OT2,T1]{fontenc}    
\usepackage{hyperref}       
\usepackage{url}            
\usepackage{booktabs}       
\usepackage{amsfonts}       
\usepackage{nicefrac}       
\usepackage{microtype}      
\usepackage{graphicx}
\usepackage{natbib}
\usepackage{doi}
\usepackage[russian,english]{babel}

\usepackage{times}
\usepackage{latexsym}
\usepackage{algorithm}
\usepackage{latexsym}
\usepackage{algpseudocode}
\usepackage{multicol}
\usepackage{amsmath}
\usepackage{enumitem}
\usepackage{multirow} 
\usepackage{algorithmicx}
\usepackage{tipa}

\usepackage[T1]{fontenc}

\title{Patterns of Closeness and Abstractness in Colexifications: \\ The Case of Indigneous Languages in the Americas}

\author{ Yiyi Chen\thanks{This work is supported by the Carlsberg Foundation under a \textit{Semper Ardens: Accelerate} career grant held by JB, entitled ``Multilingual Modelling for Resource-Poor Languages'', grant code CF21- 0454. This work is accepted to SIGMORPHON 2023. } \\
	Department of Computer Science\\
	Aalborg University\\
	Copenhagen, Denmark \\
	\texttt{yiyic@cs.aau.dk} \\
	\And
	Johannes Bjerva \\
	Department of Computer Science\\
	Aalborg University\\
	Copenhagen, Denmark \\
	\texttt{jbjerva@cs.aau.dk} 
}

\renewcommand{\shorttitle}{Patterns of Closeness and Abstractness in Colexifications}

\hypersetup{
pdftitle={A template for the arxiv style},
pdfsubject={q-bio.NC, q-bio.QM},
pdfauthor={David S.~Hippocampus, Elias D.~Striatum},
pdfkeywords={First keyword, Second keyword, More},
}

\begin{document}
\maketitle




Colexification refers to linguistic phenomena where multiple concepts (meanings) are expressed by the same lexical form, such as polysemy or homophony. Semantic typology studies this type of cross-lingual semantic categorization~\citep{evans2010semantic}. The term ``colexification" was first formalized in semantic typology by~\citet{franccois2008semantic} to create semantic maps. Colexifications have been found to be pervasive across languages and cultures. The investigation of cross-lingual colexifications has provided insights across different fields, such as psycholinguistics~\citep{jackson-2019}, cognitive science~\citep{GIBSON2019389} and linguistic typology~\citep{SchapperKoptjevskajaTamm+2022+199+209}, but remain relatively unexplored in NLP (see, e.g., \citet{harvill-etal-2022-syn2vec}, \citet{chen2023colex2lang}). The Database of Cross-Linguistic Colexifications (CLICS\textsuperscript{3})~\citep{rzymski2020database} was created to include 4,228 colexification patterns across 3,156 languages, to facilitate research in colexifications.

Certain colexifications are more prevalent than others, for example, \textsc{moon} and \textsc{month} colexify in more than 300 languages while \textsc{cat} and \textsc{dog} only colexify across 4 languages in CLICS~\textsuperscript{3}\footnote{\url{https://clics.clld.org/graphs/subgraph_1313}}. Previous works have attempted to shed light on the patterns of colexifications, e.g., why certain concepts or meanings colexify more frequently than others, and what patterns colexifications reveal. Studies suggest that related concepts are more likely to colexify~\citep{karjus2021conceptual,XU2020104280}. ~\citet{XU2020104280} argues that concepts requiring less cognitive effort to relate are more likely to colexify across languages, by investigating 246 languages and 1310 meanings using the Intercontinental Dictionary Series~\citep{borin2013intercontinental}. \textit{The Goldilocks principle} suggests that both relatedness and accurate information transfer play roles in colexifications using CLICS~\textsuperscript{3} ~\citep{goldilocks}.  For example, strongly related concepts \textsc{left} and \textsc{right} are expected to be less likely to colexify because they are hard to distinguish in context.

\begin{figure}[ht]
    \centering
    \caption{Colexification Subgraph \textsc{sun} with concreteness distance on the edge of each pair of colexified concepts. Each concept is annotated with concreteness.}
    \includegraphics[width=0.5\textwidth]{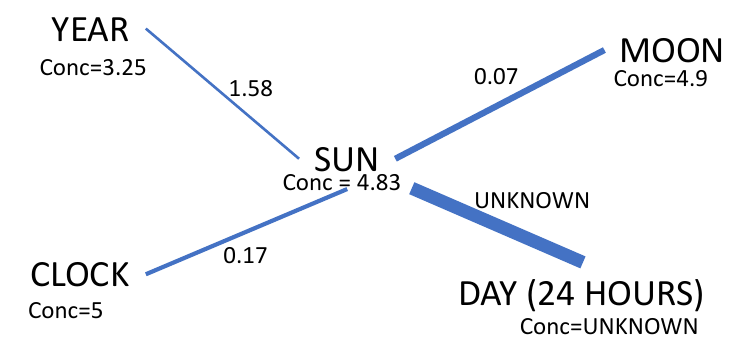}
    \label{fig:graph}
\end{figure}

 The problem of concreteness/abstractness of concepts is interdisciplinary, studied from a cognitive standpoint in linguistics, psychology, psycholinguistics, neurophysiology, etc~\citep{solovyev2021concreteness}. Concrete concepts are those that are perceived by the senses, such as \textsc{cat} and \textsc{mountain}, while abstract concepts are not perceived by the senses, such as \textsc{relationship} and \textsc{understanding}. ~\citet{brysbaert2014concreteness} curates concreteness ratings for 37,058 English words and 2,896 two-word expressions from over 4,000 participants, which has provided insights across various linguistic disciplines.
 The concreteness ratings scales from 1 (abstract) to 5 (concrete).

\begin{table*}[ht!]
    \centering
    \caption{Correlation between \#Colexifications and the Concreteness Distance between the Colexified Concepts (a) across 1989 languages in CLICS and (b) 170 Indigenous American Languages, respectively. }
    \begin{tabular}{lll||ll}
    \toprule 
    &    a && b & \\
             &  Correlation Coefficient & p-value &  Correlation Coefficient & p-value\\\midrule
      \#Colex. & -0.3843 & 4.2965e-26* &  -0.3585 & 2.0013e-19* \\
      Colex. Patterns  &-0.3500 & 1.2436e-21* &  -0.3829 & 3.8299e-22* \\  
        \#Languages & -0.3537 & 4.3349e-22* &  -0.3603 & 1.2804e-19* \\\bottomrule 
    \end{tabular}
    \label{tab:corr_americas}
\end{table*}

\begin{figure*}[ht!]
    \centering
    \caption{Histogram of Number of Colexifications against Concreteness Distance between the Colexified Concepts across 1989 languages in terms of all colexifications (left), unique collexification patterns (middle), and number of languages where colexification patterns occur (right), respectively.  }
    
    \includegraphics[width=0.32\textwidth]{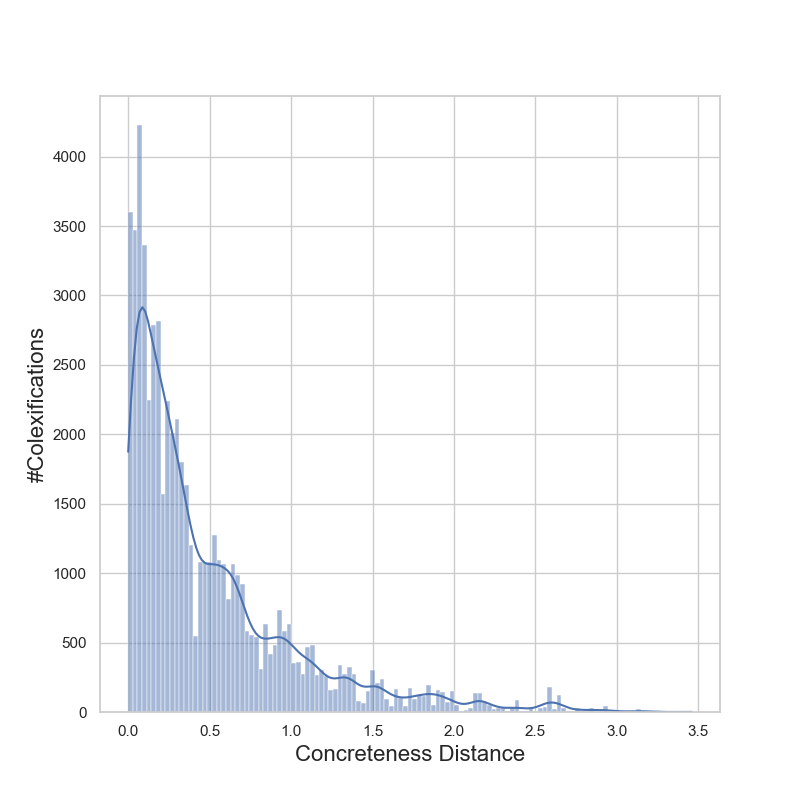}
    \includegraphics[width=0.32\textwidth]{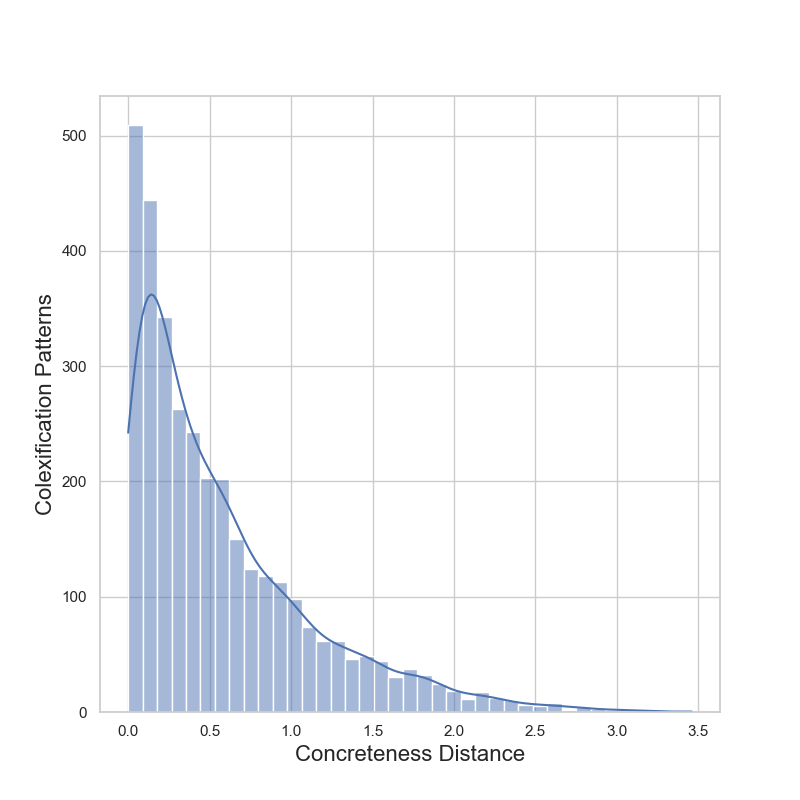}
    \includegraphics[width=0.32\textwidth]{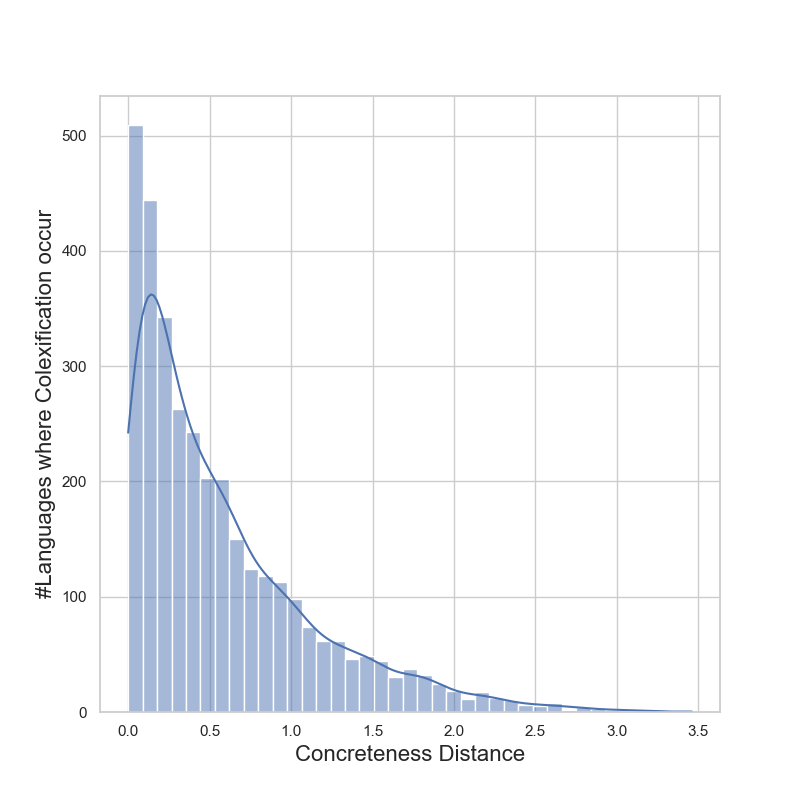}

    \label{fig:hist_corr}
\end{figure*}

Previous studies show that abstract concepts are often understood by reference to more concrete concepts~\citep{lakoff2008metaphors}, and words that first arise with concrete meanings often later gain an abstract one~\citep{xu2017evolution}.~\citet{XU2020104280} leans on these findings to show that  concepts more dissimilar in concreteness and affective valence are more likely to colexify. However, using CLICS\textsuperscript{3}, ~\citet{di2021colexification} find that colexification patterns capture similar affective meanings, i.e., the concepts that are closer in affectiveness are more likely to colexify. Moreover, studies show that abstract words are often combined with abstract ones, and vice versa for concrete words ~\citep{frassinelli-etal-2017-contextual, naumann-etal-2018-quantitative}. 

We have a markedly differing hypothesis from previous work. Unlike ~\citet{XU2020104280}, we hypothesize that concepts that are \textit{closer} in concreteness/abstractness are more likely to colexify. 
We use CLICS\textsuperscript{3} to map concepts with the concreteness list from~\citep{brysbaert2014concreteness}, comparing all languages available, with indigenous languages in the Americas. We then construct graphs of colexifications, with nodes being the concepts and annotated with concreteness, and edges weighted by languages. Fig.~\ref{fig:graph} shows an example of part of the subgraph \textsc{sun}, and the edge weights are proportional to the number of language where the respective concepts colexify. 
The concreteness distance for each colexification is calculated simply by extracting the absolute distance from one concept to another, since the colexification graph is undirected. As shown in Fig.~\ref{fig:graph}, concreteness distances of concepts are annotated on the edges, except for the edge connected to the concept \textsc{day (24 hours)} with unknown concreteness. By our hypothesis, the colexification (\textsc{year}, \textsc{sun}) with the most distant concretenesses compared to other colexifications in the graph, i.e., (\textsc{moon}, \textsc{sun}), (\textsc{clock}, \textsc{sun}) and (\textsc{day (24 hours)}, \textsc{sun}), occurs less frequently across languages. 
Fig.~\ref{fig:hist_corr} shows the histograms of the count of a) all the colexifications, b) unique colexification patterns, and c) the languages where the colexifications occur against the concreteness distances, across all 1989 languages in CLICS\textsuperscript{3}.

To test our hypothesis, we calculate the Pearson correlation  calculated for the X-axis and Y-axis presented in Fig.~\ref{fig:hist_corr}. 
As shown in Table~\ref{tab:corr_americas} there is a statistically significant negative correlation between colexifications and the concreteness distances of the colexified concepts.
This verifies our hypothesis, showing that it is indeed more likely for a pair of concepts to colexify when they have similar levels of concreteness. 


Our preliminary findings challenge previous theories and findings on the correlation between colexification and metaphoricity, corroborated by data with a broader range of concepts and languages.
We will conduct analyses to evaluate a stronger hypothesis, i.e., that colexifications encode closeness in concreteness, by leveraging language weights between concepts.
This line of research will provide insights into the study of human conceptualization, and more specifically assist in bootstrapping concreteness for large-scale lexical databases to facilitate further interdisciplinary research, such as psycholinguistics and multilingual NLP.

\section*{Limitations}
A limitation of this study is the fact that the concreteness ratings of \citet{brysbaert2014concreteness} are curated solely from self-identified U.S. residents. As such, there is a risk of an anglocentric bias in that dataset. However, as our findings are consistent both across English, languages spoken in the global West, but also generalise to, e.g., indigenous languages in the Americas, we argue that the findings of this study are robust.





\bibliographystyle{unsrtnat}
\bibliography{references}  

\end{document}